# Exploring the Potential of Large Language Models to Generate Formative Programming Feedback

Natalie Kiesler[§]
*Information Center Education*
*DIPF Leibniz Institute for Research*
*and Information in Education*
Frankfurt am Main, Germany
kiesler@dipf.de

Dominic Lohr[§]
*Computer Science Education*
*Friedrich-Alexander-University*
Erlangen, Germany
dominic.lohr@fau.de

Hieke Keuning[§]
*Information and Computing Sciences*
*Utrecht University*
Utrecht, The Netherlands
h.w.keuning@uu.nl

*Abstract*—Ever since the emergence of large language models (LLMs) and related applications, such as ChatGPT, its performance and error analysis for programming tasks have been subject to research. In this work-in-progress paper, we explore the potential of such LLMs for computing educators and learners, as we analyze the feedback it generates to a given input containing program code. In particular, we aim at (1) exploring how an LLM like ChatGPT responds to students seeking help with their introductory programming tasks, and (2) identifying feedback types in its responses. To achieve these goals, we used students' programming sequences from a dataset gathered within a CS1 course as input for ChatGPT along with questions required to elicit feedback and correct solutions. The results show that ChatGPT performs reasonably well for some of the introductory programming tasks and student errors, which means that students can potentially benefit. However, educators should provide guidance on how to use the provided feedback, as it can contain misleading information for novices.

## I. Introduction

The challenges educators have to face nowadays include highly increasing numbers of students in STEM, scarce staff resources, and heterogeneous groups of learners with their need for formative feedback [1], [2]. Hence, the educators' need for automated systems that support their teaching process is growing. Many of these systems focus on providing automated feedback, but this feedback is often still basic and not tailored to the learners' needs [3]–[5].

In late November 2022, an easily accessible tool using a large language model (LLM) was released, known as ChatGPT. The tool is designed to engage in conversations with humans and seems to provide impressive results in natural language discussions, the generation of texts, and other tasks. Despite its weaknesses, its performance and error analysis in introductory programming tasks seem promising [6], [7]. Although we do not yet know a lot about its impact, the computing education community needs to explore the potential of LLMs and related tools for teaching, learning, and assessing.

This work-in-progress paper addresses this demand by exploring the potential of an LLM-based tool for generating formative feedback on programming tasks. In particular, we investigate 99 generated responses to student input comprising solutions to introductory programming exercises and an accompanying question about the student solution. The research questions of this work are: (RQ1) How does an LLM system like ChatGPT respond to students seeking help with their introductory programming tasks? (RQ2) Which feedback types can be recognized in the responses?

## II. Related work

While LLMs have only appeared recently, several researchers have started studying their implications for education, among them an ITiCSE working group [8]. This section introduces recent work in the context of computing education specifically. One of the first papers appeared in early 2022 [6], describing the performance of OpenAI's Codex model on CS1 programming exercises, showing that it can compete well with actual students. However, the Codex model is currently not accessible anymore, because newer and better models quickly followed. The rapid replacement of models shows that studies become outdated quickly, and are hard to replicate, especially if research data is not available (anymore) [9]. A follow-up study on CS2 exercises also showed impressive results [7].

Becker et al. [10] discuss several opportunities and challenges regarding the educational opportunities and challenges of LLMs. They urge that we "need to review our educational practices in the light of these new technologies." Kiesler and Schiffner [11] further derive ChatGPT's implications for assessment. At the same time, LLMs are already applied in computing education to generate exercises and code explanations [12], [13], or to improve compiler error messages [14].

LLMs have also been used to repair buggy programs. Zhang et al. [15] applied the Codex completion model to student Python programs and found that an LLM trained on code (i.e., Codex) can fix both syntactic and semantic mistakes. Denny et al. [16] explored the engineering of prompts using GitHub Copilot. Their goal was to identify types of problems where Copilot does not perform well, and how to phrase questions in natural language to achieve good results.

Further studies directly involved students. Kazemitabaar et al. [17], for example, conducted experiments with novice programmers. Half of them had access to Codex when working on programming tasks. Their results show that student who used Codex significantly increased their code-authoring performance. Prather et al. [18] investigated students' use of GitHub Copilot in an introductory programming assignment through observation and interviews. They conclude that novices still experience challenges when using Copilot, and that Copilot's design should be improved.

[§]Equal contribution

In this study, we use a different approach, as we generate feedback with ChatGPT on student submissions to small programming problems. We then explore and categorize the generated output in detail, as opposed to simply assessing whether an AI model can fix the problem. While conducting our study, some new related works appeared [19]–[21].

## III. METHOD

### A. Data Selection

To answer the research questions, we used a dataset gathered within a CS1 course with about 300 students. Our dataset consists of student submissions to programming exercises. We selected four tasks and respective submissions to the weekly exercises. Tasks were selected based on the following criteria:

- Tasks originate from the first four weeks of the course.
- Tasks do not comprise interdependent subtasks.
- Length of solution does not exceed 15 lines of code.
- Submissions with diverse errors are available.

Based on these criteria, we selected four tasks: 1) computing thermal equations of state (TEOS), 2) classifying triangle types (TTBS and TTBA), and 3) computing the negative Fibonacci sequence (NEGF). A task description and the addressed concepts are provided in Table I.

TABLE I
EXERCISE OVERVIEW.

| Exercise | Description | Concepts |
|---|---|---|
| TEOS | Write methods for the given thermal equations of state using the constants provided. | calculations, class, methods, constants |
| TTBS | Classify a triangle by its sides (equilateral, isosceles, or scalene), depending on whether all three sides, exactly two sides, or no sides are equal. | methods, calculations, conditionals, enums |
| TTBA | Classify a triangle by its angles (acute, right, or obtuse) depending on whether all three angles are less than 90°, one angle is exactly 90°, or one angle is greater than 90°. | methods, calculations, conditionals, enums |
| NEGF | Compute the negative Fibonacci sequence for negative inputs without using classes or methods from the Java API. | methods, conditionals, recursion |

Next, we selected erroneous student solutions for each of the three tasks as input to ChatGPT, making sure that they differed in terms of error patterns. These errors could be compiler errors, logical errors as identified by test cases, and stylistic issues. Selected solutions may contain multiple errors.

### B. Data Analysis

To answer RQ1 and explore the generated feedback, the selected student solutions were used as an input to ChatGPT (March 23 Version), accompanied by the prompt "What's wrong with my code?" followed by the code of the student's submission. We analyzed ChatGPT's responses with regard to their characteristics (e.g., feedback content [22], [23]). After several iterations as part of a design-based research approach, we distinguished 11 criteria within three categories: *Content*, *Quality*, and *Other*. For each criterion, we defined a closed set of specifications (e.g., Yes/No/Snippet/Not applicable).

For the feedback's *content*, we identified elements contained in ChatGPT's responses. Among them are requests for more information (INFO), stylistic suggestions (STYLE), textual explanations of the cause of an error (CAUSE), its fix (FIX), and whether code (CODE) or examples (EXA) are provided. Moreover, we analyzed the *quality* of ChatGPT's output w.r.t. code compilation (COMP), misleading information (MIS) and uncertainty (UNC). *Other* criteria comprise meta-cognitive (META) and motivational elements (MOT) (see Table III).

For the exploration of how the feedback deviates upon regeneration, we generated three responses for each submission (using the "regenerate"-option), by one author, on the same machine, in the same browser, and in one chat per student. We assume students to act similarly when using ChatGPT. Each of the 99 outputs generated by ChatGPT was analyzed by two authors. Disagreements were discussed and resolved using a consensual approach.

To answer RQ2 relating to feedback types, we refer to an existing typology of elaborated feedback in the context of programming education [5], based on Narciss' feedback classification [23]. Hence, ChatGPT's responses to the prompt are analyzed with regard to the following five types of feedback (subtypes not mentioned here for brevity):

- KTC: Knowledge about task constraints
- KC: Knowledge about concepts
- KM: Knowledge about mistakes
- KH: Knowledge on how to proceed
- KMC: Knowledge about meta-cognition

## IV. RESULTS

In this section, we present the results of the analysis as answers to the two research questions of this work.

### A. How ChatGPT responds to student input (RQ1)

To answer the first research question, we explored the characteristics of ChatGPT's output. In Table II, all of the aforementioned criteria (see Section III-B) are presented along with a shorthand notation representing the degree to which each criterion is fulfilled. Moreover, the table contains the results for each of the three generations requested by the authors. This is why we display three columns with results for each criterion (labeled as R1, R2, and R3). Table III explains all abbreviations used for the characteristics.

First of all, we recognized that the three generated responses from ChatGPT to the same student input vary greatly. The 11

TABLE II
HOW CHATGPT RESPONDS TO STUDENT INPUT (RQ1). FOR LEGEND SEE TABLE III

| Stud_task | CONTENT | | | | | | | | | | | | | | | | | | QUALITY | | | | | | | | | OTHER | | | | | |
|---|---|---|---|---|---|---|---|---|---|---|---|---|---|---|---|---|---|---|---|---|---|---|---|---|---|---|---|---|---|---|---|---|---|
| | INFO | | | STYLE | | | CAUSE | | | FIX | | | CODE | | | EXA | | | COMP | | | MIS | | | UNC | | | META | | | MOT | | |
| | R1 | R2 | R3 | R1 | R2 | R3 | R1 | R2 | R3 | R1 | R2 | R3 | R1 | R2 | R3 | R1 | R2 | R3 | R1 | R2 | R3 | R1 | R2 | R3 | R1 | R2 | R3 | R1 | R2 | R3 | R1 | R2 | R3 |

(●: Yes  ○: No  ◐: Snippet  –: Not applicable)

TABLE III
LEGEND FOR CODING.

| Symbol | Meaning |
|---|---|
| INFO | Requesting more information |
| STYLE | Stylistic suggestion |
| CAUSE | Textual explanation cause of error |
| FIX | Textual explanation fix of error |
| CODE | Code provided |
| EXA | Illustrating examples |
| COMP | Code, if provided, compiles |
| MIS | Misleading information |
| UNC | Uncertainty |
| META | Meta-cognitive elements |
| MOT | Motivational elements |

criteria, which are applicable to all 33 tasks, show variations in 116 out of the 363 triples, appearing in at least one of the three regenerated replies. Due to this degree of randomness, it seems challenging to rely on responses or to know what to expect as an answer.

ChatGPT's responses usually contained textual explanations of the cause and fix of errors (88, and 83 responses each). In 65 out of 99 responses, an improved version of the code was offered, in 13 other cases, at least code snippets were provided. Stylistic suggestions (e.g., on comments, naming conventions) were less common, with 30 occurrences. In a few cases (4 out of 99) ChatGPT also requested more information about the problem (e.g., *"If you could provide more context about what you are trying to achieve and any error messages you are encountering, I can help you more effectively."*).

In 61 of the 99 generated responses, ChatGPT offered misleading information to the asking person, meaning one or more aspects within the answer were incorrect or would not have contributed to improving the solution, especially for novices. ChatGPT further uttered uncertainty in 21 of the 99 generated responses across all four tasks (e.g., *"Based on the code you provided, it is difficult to identify any errors"*). Most instances (14) were noticed in responses to the NEGF task.

Moreover, we noted a few cases of meta-cognitive (e.g., knowledge about knowledge) and encouraging elements (e.g., *"I hope this helps!"*) in the responses. The meta-cognitive aspects were related to strategic knowledge and, for example, how to test the solution.

### B. Types of feedback provided by ChatGPT (RQ2)

As the next step, we analyzed ChatGPT's output w.r.t. the existing feedback typology [5]. We noted correlations between the inductively built characteristics and the well-known feedback types during this process.

For example, ChatGPT did not provide any feedback within the category of KTC feedback. This is not surprising, as the LLM cannot consider task constraints unless one additionally enters the task description. For the other four main types of feedback, we identified corresponding categories. ChatGPT provides KC feedback via examples, denoted as EXA within the content category. KM feedback is also evident within the CAUSE or STYLE elements provided by ChatGPT, as these categories provide explanations about the mistake. KH feedback corresponds to the categories FIX and STYLE, as ChatGPT expresses how to fix errors or improve results. The last main type of KMC is somewhat reflected within the META category of our characterization scheme.

In addition, we noted that ChatGPT provides knowledge about correct response feedback (KCR), which is one of the simple feedback types [5], [23]. This is particularly evident when the responses contain the full, improved code.

Another aspect worth mentioning is the lack of quality aspects in the used feedback typology [5]. Moreover, the LLM's request for more information or the notion of uncertainty are entirely new categories. In the present analysis, it was crucial to evaluate the feedback's quality, as it was generated, and not developed by a human. Even though many established feedback types were applicable in our analysis of ChatGPT's responses, it may be useful to expand the feedback typology.

## V. DISCUSSION AND LIMITATIONS

The results described in this paper show that LLMs like ChatGPT have the potential to address the discrepancy between learners' need for formative feedback on the root causes of an error [24]. Textual explanation of errors and fixes along with improved code are key characteristics of ChatGPT's feedback compared to other learning environments. The majority of recent tools provide simple feedback, report failed test cases, or compiler errors [4], [5].

At the same time, the results raise concerns about the reliability and suitability of LLM-generated feedback. As the characteristics of the output greatly vary and contain misleading information in many cases, it is questionable whether this is suitable for novice programmers. They would need the competency to formulate appropriate prompts. Due to the random nature of LLM-generated feedback, learners will receive different feedback for the same input. Furthermore, novice programmers may lack the competency to critically analyze feedback and consider all task constraints, as they do not yet have acquired competencies within the meta-cognitive dimension [25]–[27].

From our analysis, it remains open whether and how additional input to the LLM would reduce the share of misleading feedback. For example, some of the student submissions included package imports, which are prohibited in the context of the lecture. In the TEOS task, constants were provided in an outsourced file. Since ChatGPT was not provided with this information in the prompt, errors due to incorrect imports or the incorrect use of constants were not addressed and caused some of the misleading information.

Furthermore, the selection of suitable tasks and submissions from the data proved to be challenging. Most tasks were divided into several interdependent subtasks. Therefore, only a few tasks could be isolated for this analysis. Additionally, many of the submissions contained a large number of different errors, making it difficult to build a diverse dataset without redundancies.

The results presented in this study are limited to the observations made using a single LLM trained on general data. In the meantime, a new version of ChatGPT (Version 4) and other LLMs – specifically trained on programming data – have been published, which may have improved performance compared to the model used in this study. Moreover, we only explored the characteristics of the feedback based on four tasks and available student solutions from one introductory programming course. The generated feedback is likely to vary for other tasks. The same is true for different prompts. So far, we only explored one general, very open prompt, as we assume students to behave similarly.

## VI. CONCLUSION AND NEXT STEPS

In this work-in-progress, we explored the potential of Large Language Models such as ChatGPT to generate formative feedback to novice learners of programming. To achieve this, we investigated ChatGPT's responses to incorrect student solutions to introductory programming tasks. As part of the analysis, we characterized ChatGPT's output w.r.t. its content, quality and other criteria. Moreover, we applied the feedback typology by Keuning et al. [5] to the responses. The results show that an LLM like ChatGPT performs reasonably well when it comes to the detection and correction of compile errors. The availability of textual explanations and improved code (snippets) aligned to the user's input is also unique. However, there are limitations to ChatGPT's feedback quality on logic and semantic errors, or if multiple errors are contained in a student solution. Moreover, ChatGPT may provide misleading information, and it lacks information on task constraints. Hence, it is crucial to guide students towards using ChatGPT, to inform them about its capabilities and limitations, and to develop respective pedagogical methods.

As the next step of this work-in-progress, we will evaluate the quality of feedback generated by LLM-based tools by comparing the generated output with expert feedback on these tasks. Future work will also comprise an investigation of how we can engineer prompts to generate correct solutions from a student input, and to generate specific types of feedback. Based on such research, we can further discuss the implications for computing educators considering the use of LLMs in the classroom, and how we can adapt pedagogy to the availability of LLMs. Another continuation of this work is the combination of the presented feedback characterization with the existing feedback typology for the context of programming. This is how we can develop a more recent framework for the classification of feedback types in the robot age.